\documentclass[a4paper,conference]{IEEEtran}
\ifCLASSINFOpdf
  \usepackage[pdftex]{graphicx}
\else
\fi
\usepackage{amsmath}
\begin{document}
%
\title{Minority Class Oversampling for Tabular Data with Deep Generative Models}

\author{\IEEEauthorblockN{Ramiro D. Camino, Radu State}
\IEEEauthorblockA{SnT, University of Luxembourg, Luxembourg\\
\{ramiro.camino,radu.state\}@uni.lu}
\and
\IEEEauthorblockN{Christian A. Hammerschmidt}
\IEEEauthorblockA{Delft University of Technology, Netherlands \\
c.a.hammerschmidt@tudelft.nl}
}


%


\maketitle

\begin{abstract}
In practice, machine learning experts are often confronted with imbalanced data.
Without accounting for the imbalance, common classifiers perform poorly and standard evaluation metrics mislead the practitioners on the model's performance.
A common method to treat imbalanced datasets is under- and oversampling. 
In this process, samples are either removed from the majority class or synthetic samples are added to the minority class.
In this paper, we follow up on recent developments in deep learning.
We take proposals of deep generative models, including our own, and study the ability of these approaches to provide realistic samples that improve performance on imbalanced classification tasks via oversampling.

Across 160K+ experiments, we show that all of the new methods tend to perform better than simple baseline methods such as SMOTE, but require different under- and oversampling ratios to do so.
Our experiments show that the way the method of sampling does not affect quality, but runtime varies widely.
We also observe that the improvements in terms of performance metric, while shown to be significant when ranking the methods, often are minor in absolute terms, especially compared to the required effort.
Furthermore, we notice that a large part of the improvement is due to undersampling, not oversampling.
We make our code and testing framework available.

\end{abstract}

%
\IEEEpeerreviewmaketitle

\section{Introduction}
\label{sec:introduction}

With recent advances in the field of generative adversarial networks (GANs)~\cite{goodfellow_generative_2014}, such as  learning to transfer properties~\cite{kim_learning_2017}, advances in practical aspects \cite{gulrajani_improved_2017}, or advances in understanding theoretical aspects \cite{mescheder_which_2018}, it is tempting to apply GANs as tools to problems in data science tasks.
Data available in real-word settings often has quality issues, like class imbalances and the lack of ground truth.
Generative methods learning the distribution of the data using deep learning, such as GANs and variational autoencoders (VAEs)\cite{kingma_auto-encoding_2013}, can help building solutions.
While deep generative models (DGM) work well on continuous-domain problems such as images and video, they struggle generating samples in the form of discrete sequence and categorical or mixed distributions. 
This is in part due to the inherent difficulty of training networks with discrete outputs: sampling from discrete distributions is a non-differentiable operation, which makes it impossible to train the network using backpropagation.
To use DGMs for imputation, simulation, feature extraction, transfer learning, or sampling artificial datapoints these limitations need to be addressed.

In the context of data science and in business-oriented applications, data is often tabular, i.e. contains multiple categorical and numerical variables.
Discrete variables are also important in natural language modeling problems and reinforcement learning tasks.
While initial proposals focused on continuous variables, following proposals addressed  generating multivariate binary samples~\cite{choi_generating_2017} or sequential samples from a single categorical variable~\cite{kusner_gans_2016,yu_seqgan:_2017,junbo_adversarially_2017,gulrajani_improved_2017,camino2018generating,camino2019improving}.

Based on these methods, several papers addressed the issue of generating samples over distributions of discrete or tabular data, including generating airline data~\cite{mottini2018airline} or medical time series ~\cite{choi2017generating} as well as frameworks to solve tasks such as oversampling~\cite{douzas2018effective} or multiple dependent tasks such as HexaGAN~\cite{hwang2019hexagan}.
Yet, comparing the performance of different architectures on a single task is difficult:
Besides the use of different datasets and multiple setups for hyperparameter search, the protocols to solve tasks, such as the under- and oversampling ratios, vary from proposal to proposal.

We enumerate our contributions as follows:
\begin{itemize}
\item We present a short review of recent approaches to learning categorical distributions with deep networks,
\item we (re-)implement generative models and extend them for multivariate situations where necessary and make the implementations available, and
\item we conduct a large-scale evaluation of the model performance with fixed model-parameters, a clear protocol, and a fair hyperparameter search, and a comparative study on different methods of sampling from the models.
\end{itemize}

In the following sections, we present related work (Section \ref{sec:related}), our protocol and comparision approach (Section \ref{sec:comparison}), the experiments (Section \ref{sec:experiments}) and our conclusions (Section \ref{sec:conclusion}).
\section{Related Work}
\label{sec:related}

Unsupervised learning from datasets is of interest in many fields, serving different purposes.
Examples include (Bayesian) inference tasks as in~\cite{mansinghka_bayesdb:_2015}, where the authors learn Bayesian models using non-parametric priors over multivariate tabular data, as well as unsupervised feature discovery, extraction, and transfer~\cite{kim_learning_2017,radford_unsupervised_2015,isola_image--image_2016}, and synthetic data for privacy aware applications \cite{choi_generating_2017}.
In the context of imbalanced datasets, SMOTE~\cite{chawla_smote:_2002} is a well-known method for generating synthetic samples which combined with an over- and undersampling protocol can improve classifier performance.
Several variants of SMOTE were proposed~\cite{more_survey_2016}, each addressing different shortcomings or extending the applicability of the method to other types of datasets and variable distributions.

On continuous data, GANs~\cite{goodfellow_generative_2014} have proven to be good at learning data distributions in an unsupervised fashion. 
By feeding in additional information (such as labels), conditional GANs~\cite{mirza_conditional_2014} can learn to generate samples for specific inputs.
On discrete data (e.g. text sequences, categorical data), GANs face problems leading to a comparatively worse performance.
To deal with discrete data, several approaches exist. The Gumbel-Softmax \cite{jang_categorical_2016} and the Concrete-Distribution \cite{maddison_concrete_2016} were simultaneously proposed to tackle this problem in the domain of variational autoencoders (VAE) \cite{kingma_auto-encoding_2013}.
Later, \cite{kusner_gans_2016} adapted the technique to GANs for sequences of discrete elements.

Addressing the same problem, a reinforcement learning approach called SeqGAN \cite{yu_seqgan:_2017} interprets the generator as a stochastic policy and performs gradient policy updates to avoid the problem of backpropagation with discrete sequences.
The discriminator outputs the reinforcement learning reward for a full sequence, and several simulations generated with Monte Carlo search are executed to complete the missing steps.

Adversarially Regularized Autoencoders (ARAE) \cite{junbo_adversarially_2017} transform sequences from a discrete vocabulary into a continuous latent space while simultaneously training both a generator (to output samples from the same latent distribution) and a discriminator (to distinguish between real and fake latent codes).
The approach relies on Wasserstein GAN (WGAN) \cite{arjovsky_wasserstein_2017} to improve training stability and obtain a loss more correlated with sample quality.
The main two changes in WGAN consist on replacing the discriminator by a critic, and limiting the size of the critic parameters by a constant.

MedGAN \cite{choi_generating_2017}, while architecturally similar, is used to synthesize realistic health care patient records.
The method pre-trains an autoencoder and then the generator returns latent codes as in the previous case, but they pass that output to the decoder before sending it to the discriminator; therefore, the discriminator receives either fake or real samples directly instead of continuous codes.
They propose using shortcut connections in the generator.
Additionally, the authors present a technique called mini-batch averaging in order to better evaluate the generation of a whole batch instead of individual isolated samples.
Before feeding a batch into the discriminator, mini-batch averaging appends the average value per dimension of the batch to the batch itself.

To address the difficulty of training GANs, an improved version for WGAN is presented in \cite{gulrajani_improved_2017} adding a gradient penalty to the critic loss (WGAN-GP) and removing the size limitation of the critic parameters.
The authors present a use case for the generation of word sequences, where they claim that during training, discrete samples can be generated just by passing the outputs of softmax layers from the generator to the discriminator without sampling from those outputs.
\section{Comparison}
\label{sec:comparison}

\subsection{Datasets}

Datasets like MNIST, ImageNet and CIFAR are very well known in the domain of computer vision.
Thanks to this, countless studies could compare their findings against others using widely accepted benchmarks.
On the other side, when applying deep learning to tabular data, there is no framework that is that well defined.
Several papers opted to work with datasets from the UCI Repository~\cite{dua_uci_2017} on tasks related to this domain: tabular data imputation~\cite{yoon2018gain,nazabal2018handling,gondara2018mida,mattei2018miwae}, imbalanced classification using a latent space~\cite{ng2016dual}, oversampling from deep generative models~\cite{douzas2018effective,xu2019modeling} or all the previous tasks at the same time~\cite{hwang2019hexagan}.
Nevertheless, we want to point out several interesting aspects about the datasets selected across these studies.
First of all, the datasets usually are presented with short names or aliases which can lead to confusion.
For example, one dataset referred as ``breast'' or ``breast-cancer'' presents four versions online with different features, and sometimes it is not very clear which one the study is referring to.
Second, some datasets contain less than a thousand samples or just a few of features.
While this can be reasonable for other machine learning models, deep learning models may not reach their full potential when training with datasets of such dimensions.
And third, most of the classification tasks associated with imbalanced UCI datasets can be easily solved with state-of-the-art machine learning algorithms like XGBoost~\cite{Chen:2016:XST:2939672.2939785} without much need of any additional treatment.
For example, one study~\cite{ng2016dual} showed that 14 datasets from the UCI Repository can be classified with $ROCAUC > 0.9$ (and for most of the cases very close to 1) with a simple Random Forest~\cite{liaw2002classification}.
Before starting with this study, we used XGBoost to classify the 10 most used datasets from UCI repository among the related work.
Some of them are multi-class problems, so we transformed them into several one-vs-all binary classification problems, as well as several one-vs-one binary classification problems by pairing two classes at a time.
We found only two cases where the test f1 score was less than 0.95: ``Adult''\footnote{http://archive.ics.uci.edu/ml/datasets/adult} and ``Default of Credit Card Clients''\footnote{https://archive.ics.uci.edu/ml/datasets/default+of+credit+card+clients}.

\subsection{Deep Generative Models}

\begin{figure*}[!t]
\centering
\begin{minipage}{.5\textwidth}
  \centering
  \includegraphics[width=\columnwidth]{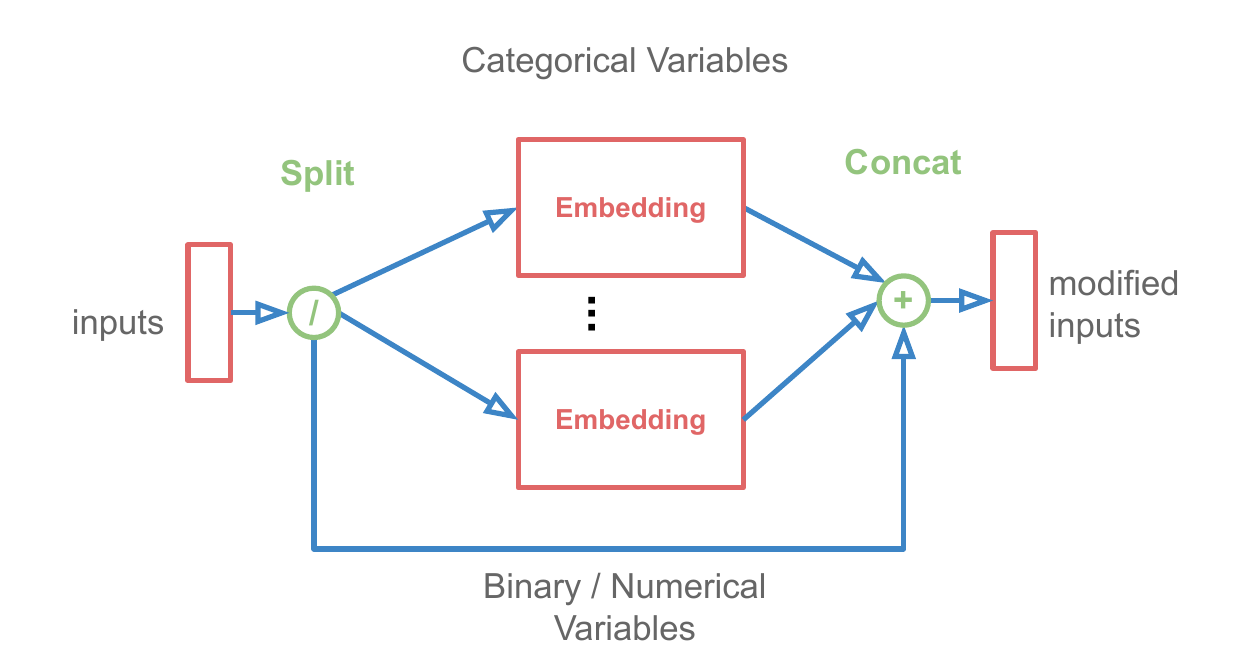}
\end{minipage}%
\begin{minipage}{.5\textwidth}
  \centering
  \includegraphics[width=\columnwidth]{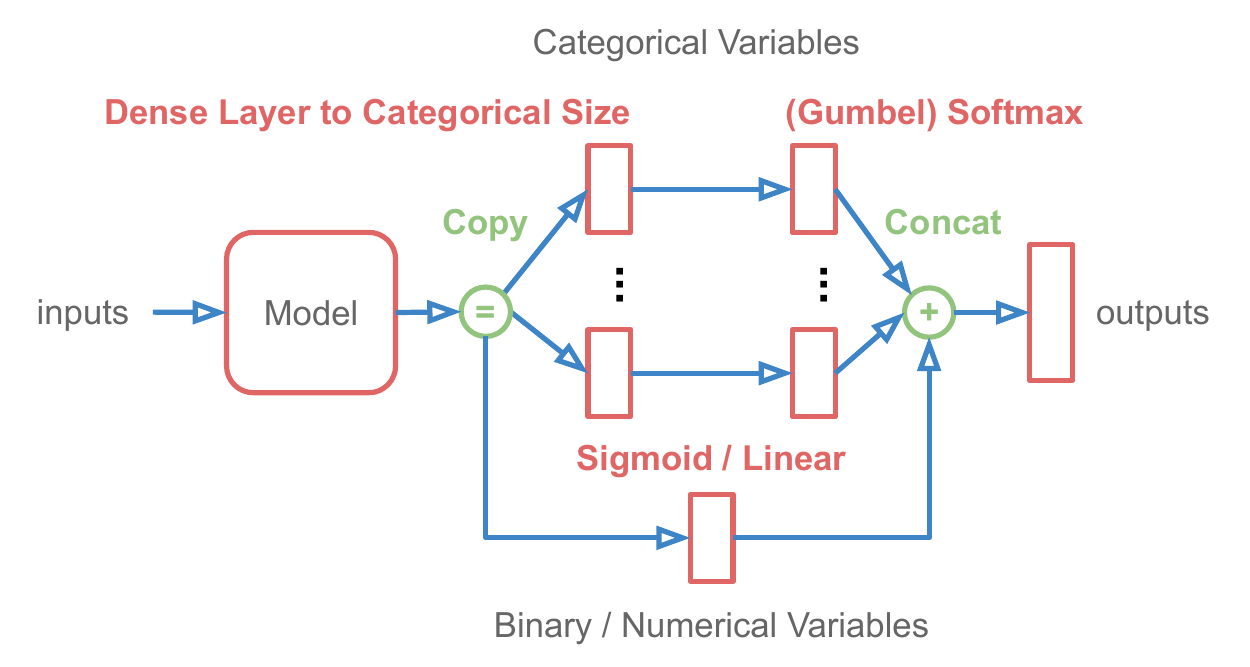}
\end{minipage}
\caption{Architecture for multi-input (left) and multi-output variables (right).} 
\label{fig:multi-variables}
\end{figure*}

In this study, we compare two well known deep generative architectures: GAN~\cite{goodfellow_generative_2014} and VAE~\cite{kingma_auto-encoding_2013}.
We include as well two variants of GAN that involve autoencoders, ARAE~\cite{junbo_adversarially_2017} and MedGAN~\cite{choi_generating_2017}, adapting the reconstruction loss by separating the features per variable (e.g. all the features from a one-hot-encoded categorical variable).
Each separated reconstruction loss depend on the type of the variable: cross entropy for categorical variables, binary cross entropy for binary variables, and mean squared error for numerical variables.
This is based on a previous work using GAN only with categorical variables~\cite{camino2018generating}, and a more complex version can be found in HI-VAE~\cite{nazabal2018handling}.
We further include Multi-Variable~\cite{camino2018generating,camino2019improving} (MV) versions of the previous models.
In Fig.~\ref{fig:multi-variables} we show how the inputs and the outputs of models are separated by variable, where categorical variables are treated in a different way than the other variables.
We selected gumbel-softmax~\cite{jang_categorical_2016,maddison_concrete_2016} for the categorical activations.
For all the Multi-Variable versions of GAN we use instead WGAN~\cite{arjovsky_wasserstein_2017}, and we also add an additional alternative with WGAN-GP~\cite{gulrajani_improved_2017}.

\subsection{Sampling from Deep Generative Models}

Among the literature we found two alternatives to generate synthetic samples with deep generative models for later use in imbalanced classification tasks.
The authors in~\cite{fiore2019using} train a GAN only using the minority class samples in order improve the detection of credit card frauds.
We call this technique ``minority''.
The second method~\cite{douzas2018effective} uses a GAN with a condition (the label) as an additional input for both the generator and the discriminator.
We will refer to this alternative as ``conditional''.
We propose an additional technique that we call ``rejection'', that consists on training any deep generative model attaching the label to the rest of the features as an additional variable, and afterwards during the synthetic generation, all the samples that do not belong to a desired class are discarded.
Note that for obtaining a synthetic sample of a specific size with this technique, iterative draws are necessary, and the procedure could never end.
Therefore, a limit on the number of draws or the execution time needs to be implemented.
\section{Experiments}
\label{sec:experiments}

In this section we provide the details and the analysis of our experiments.
The goal is to provide an empirical comparison of how different undersampling and oversampling techniques affect the classification of imbalanced datasets.
The code for classification, undersampling, oversampling and all the tasks related to deep generative models can be found online~\footnote{https://github.com/rcamino/deep-generative-models}.

\begin{table*}[ht]
\caption{Classification experiments with XGBoost for the ``Credit card fraud'' dataset.}\label{table:creditcardfraud}
\begin{center}
\begin{tabular}{|l|l|c|c|c|c|}
\hline
\multicolumn{2}{|l|}{\bfseries Technique} & \multicolumn{2}{c|}{\bfseries IR} & \textbf{Train f1} & \textbf{Test f1} \\
\hline
\multicolumn{2}{|l|}{Only classifier} & \multicolumn{2}{c|}{$0.001$} & $0.904 \pm 0.005$ & $0.814 \pm 0.046$ \\
\hline
\multicolumn{2}{|l|}{\bfseries Technique} & \multicolumn{2}{c|}{\bfseries USR} & \textbf{Train f1} & \textbf{Test f1} \\
\hline
\multicolumn{2}{|l|}{Undersampling and classifier} & \multicolumn{2}{c|}{$0.007$} & $0.861 \pm 0.009$ & $0.813 \pm 0.057$ \\
\hline
\multicolumn{2}{|l|}{\bfseries Oversampling} & \textbf{USR} & \textbf{OSR} & \textbf{Train f1} & \textbf{Test f1} \\
\hline
\multicolumn{2}{|l|}{BorderlineSMOTE} & $0.004$ & $0.005$ & $0.877 \pm 0.010$ & $0.811 \pm 0.055$ \\
\multicolumn{2}{|l|}{RandomOverSampler} & $0.002$ & $0.007$ & $0.895 \pm 0.006$ & $0.822 \pm 0.052$ \\
\multicolumn{2}{|l|}{SMOTE} & $0.002$ & $0.007$ & $0.891 \pm 0.006$ & $0.831 \pm 0.043$ \\
\multicolumn{2}{|l|}{SVMSMOTE} & $0.002$ & $0.003$ & $0.899 \pm 0.006$ & $0.816 \pm 0.062$ \\
\hline
\textbf{DGM} & \textbf{Sampling} & \textbf{USR} & \textbf{OSR} & \textbf{Train f1} & \textbf{Test f1} \\
\hline
vae & Minority & $0.002$ & $0.009$ & $0.905 \pm 0.006$ & $0.820 \pm 0.039$ \\
arae & Minority & $0.004$ & $0.005$ & $0.873 \pm 0.014$ & $0.814 \pm 0.052$ \\
medgan & Minority & $0.002$ & $0.010$ & $0.896 \pm 0.006$ & $0.822 \pm 0.042$ \\
gan & Minority & $0.006$ & $0.008$ & $0.860 \pm 0.013$ & $0.820 \pm 0.069$ \\
\hline
vae & Conditional & $0.006$ & $0.009$ & $0.871 \pm 0.008$ & $0.817 \pm 0.062$ \\
arae & Conditional & $0.002$ & $0.007$ & $0.902 \pm 0.004$ & $0.816 \pm 0.048$ \\
medgan & Conditional & $0.003$ & $0.007$ & $0.886 \pm 0.007$ & $0.810 \pm 0.056$ \\
gan & Conditional & $0.004$ & $0.010$ & $0.877 \pm 0.011$ & $0.815 \pm 0.051$ \\
\hline
vae & Rejection & \multicolumn{4}{c|}{\textit{Timeout}} \\
arae & Rejection & \multicolumn{4}{c|}{\textit{Timeout}} \\
medgan & Rejection & \multicolumn{4}{c|}{\textit{Timeout}} \\
gan & Rejection & \multicolumn{4}{c|}{\textit{Timeout}} \\
\hline
\end{tabular}
\end{center}
\end{table*}
\begin{table*}[!t]
\renewcommand{\arraystretch}{1.3}
\caption{Classification experiments with XGBoost for the ``Adult'' dataset.}
\label{table:adult}
\centering
\begin{tabular}{|l|l|c|c|c|c|}
\hline
\multicolumn{2}{|l|}{\bfseries Technique} & \multicolumn{2}{c|}{\bfseries IR} & \textbf{Train f1} & \textbf{Test f1} \\
\hline
\multicolumn{2}{|l|}{Only classifier} & \multicolumn{2}{c|}{$0.33$} & $0.743 \pm 0.002$ & $0.716 \pm 0.005$ \\
\hline
\multicolumn{2}{|l|}{\bfseries Technique} & \multicolumn{2}{c|}{\bfseries USR} & \textbf{Train f1} & \textbf{Test f1} \\
\hline
\multicolumn{2}{|l|}{Undersampling and classifier} & \multicolumn{2}{c|}{$0.50$} & $0.756 \pm 0.002$ & $0.731 \pm 0.010$ \\
\hline
\multicolumn{2}{|l|}{\bfseries Oversampling} & \textbf{USR} & \textbf{OSR} & \textbf{Train f1} & \textbf{Test f1} \\
\hline
\multicolumn{2}{|l|}{ADASYN} & $0.4$ & $0.6$ & $0.749 \pm 0.001$ & $0.727 \pm 0.010$ \\
\multicolumn{2}{|l|}{BorderlineSMOTE} & $0.5$ & $0.6$ & $0.751 \pm 0.001$ & $0.730 \pm 0.008$ \\
\multicolumn{2}{|l|}{KMeansSMOTE} & $0.5$ & $0.6$ & $0.752 \pm 0.002$ & $0.731 \pm 0.009$ \\
\multicolumn{2}{|l|}{RandomOverSampler} & $0.5$ & $0.6$ & $0.756 \pm 0.002$ & $0.732 \pm 0.007$ \\
\multicolumn{2}{|l|}{SMOTE} & $0.5$ & $0.6$ & $0.752 \pm 0.002$ & $0.732 \pm 0.008$ \\
\multicolumn{2}{|l|}{SMOTENC} & $0.6$ & $0.7$ & $0.662 \pm 0.001$ & $0.642 \pm 0.010$ \\
\multicolumn{2}{|l|}{SVMSMOTE} & $0.5$ & $0.6$ & $0.750 \pm 0.001$ & $0.730 \pm 0.009$ \\
\hline
\textbf{DGM} & \textbf{Sampling} & \textbf{USR} & \textbf{OSR} & \textbf{Train f1} & \textbf{Test f1} \\
\hline
vae & Minority & $0.5$ & $1.0$ & $0.756 \pm 0.001$ & $0.732 \pm 0.010$ \\
mv-vae & Minority & $0.5$ & $1.0$ & $0.755 \pm 0.002$ & $0.733 \pm 0.010$ \\
arae & Minority & $0.5$ & $1.0$ & $0.755 \pm 0.001$ & $0.733 \pm 0.010$ \\
mv-arae & Minority & $0.5$ & $1.0$ & $0.752 \pm 0.001$ & $0.732 \pm 0.009$ \\
medgan & Minority & $0.5$ & $0.6$ & $0.755 \pm 0.002$ & $0.732 \pm 0.007$ \\
mv-medgan & Minority & $0.6$ & $0.8$ & $0.752 \pm 0.001$ & $0.733 \pm 0.010$ \\
gan & Minority & $0.6$ & $0.7$ & $0.754 \pm 0.001$ & $0.734 \pm 0.010$ \\
mv-wgan & Minority & $0.5$ & $1.0$ & $0.752 \pm 0.002$ & $0.734 \pm 0.008$ \\
mv-wgan-gp & Minority & $0.6$ & $0.7$ & $0.754 \pm 0.001$ & $0.731 \pm 0.009$ \\
\hline
vae & Conditional & $0.6$ & $1.0$ & $0.754 \pm 0.002$ & $0.732 \pm 0.008$ \\
mv-vae & Conditional & $0.5$ & $0.6$ & $0.755 \pm 0.001$ & $0.733 \pm 0.007$ \\
arae & Conditional & $0.5$ & $0.9$ & $0.755 \pm 0.002$ & $0.734 \pm 0.009$ \\
mv-arae & Conditional & $0.6$ & $0.9$ & $0.752 \pm 0.002$ & $0.732 \pm 0.010$ \\
medgan & Conditional & $0.5$ & $1.0$ & $0.754 \pm 0.002$ & $0.733 \pm 0.008$ \\
mv-medgan & Conditional & $0.5$ & $0.8$ & $0.753 \pm 0.002$ & $0.732 \pm 0.008$ \\
gan & Conditional & $0.6$ & $0.8$ & $0.755 \pm 0.002$ & $0.733 \pm 0.011$ \\
mv-wgan & Conditional & $0.5$ & $0.6$ & $0.752 \pm 0.002$ & $0.733 \pm 0.005$ \\
mv-wgan-gp & Conditional & $0.6$ & $0.8$ & $0.752 \pm 0.003$ & $0.733 \pm 0.008$ \\
\hline
vae & Rejection & $0.5$ & $1.0$ & $0.756 \pm 0.002$ & $0.732 \pm 0.008$ \\
mv-vae & Rejection & $0.7$ & $1.0$ & $0.751 \pm 0.002$ & $0.732 \pm 0.009$ \\
arae & Rejection & $0.5$ & $1.0$ & $0.755 \pm 0.002$ & $0.733 \pm 0.007$ \\
mv-arae & Rejection & $0.6$ & $0.9$ & $0.753 \pm 0.002$ & $0.732 \pm 0.011$ \\
medgan & Rejection & $0.6$ & $0.9$ & $0.753 \pm 0.002$ & $0.733 \pm 0.010$ \\
mv-medgan & Rejection & $0.5$ & $1.0$ & $0.753 \pm 0.001$ & $0.732 \pm 0.008$ \\
gan & Rejection & \multicolumn{4}{c|}{\textit{Timeout}} \\
mv-wgan & Rejection & $0.5$ & $0.7$ & $0.754 \pm 0.002$ & $0.732 \pm 0.007$ \\
mv-wgan-gp & Rejection & $0.5$ & $0.9$ & $0.754 \pm 0.001$ & $0.732 \pm 0.008$ \\
\hline
\end{tabular}
\end{table*}
\begin{table*}[ht]
\caption{Classification experiments with XGBoost for the ``Default of credit card clients'' dataset.}
\label{table:default-of-credit-card-clients}
\centering
\begin{tabular}{|l|l|c|c|c|c|}
\hline
\multicolumn{2}{|l|}{\bfseries Technique} & \multicolumn{2}{c|}{\bfseries IR} & \textbf{Train f1} & \textbf{Test f1} \\
\hline
\multicolumn{2}{|l|}{Only classifier} & \multicolumn{2}{c|}{$0.28$} & $0.479 \pm 0.006$ & $0.457 \pm 0.036$ \\
\hline
\multicolumn{2}{|l|}{\bfseries Technique} & \multicolumn{2}{c|}{\bfseries USR} & \textbf{Train f1} & \textbf{Test f1} \\
\hline
\multicolumn{2}{|l|}{Undersampling and classifier} & \multicolumn{2}{c|}{$0.80$} & $0.552 \pm 0.004$ & $0.534 \pm 0.031$ \\
\hline
\multicolumn{2}{|l|}{\bfseries Oversampling} & \textbf{USR} & \textbf{OSR} & \textbf{Train f1} & \textbf{Test f1} \\
\hline
\multicolumn{2}{|l|}{ADASYN} & $0.8$ & $1.0$ & $0.552 \pm 0.004$ & $0.537 \pm 0.028$ \\
\multicolumn{2}{|l|}{BorderlineSMOTE} & $0.6$ & $0.7$ & $0.550 \pm 0.003$ & $0.535 \pm 0.027$ \\
\multicolumn{2}{|l|}{KMeansSMOTE} & $0.7$ & $1.0$ & $0.549 \pm 0.004$ & $0.537 \pm 0.029$ \\
\multicolumn{2}{|l|}{RandomOverSampler} & $0.8$ & $0.9$ & $0.553 \pm 0.005$ & $0.538 \pm 0.031$ \\
\multicolumn{2}{|l|}{SMOTE} & $0.6$ & $0.8$ & $0.550 \pm 0.004$ & $0.535 \pm 0.031$ \\
\multicolumn{2}{|l|}{SMOTENC} & $0.9$ & $1.0$ & $0.488 \pm 0.003$ & $0.467 \pm 0.029$ \\
\multicolumn{2}{|l|}{SVMSMOTE} & $0.8$ & $0.9$ & $0.549 \pm 0.004$ & $0.534 \pm 0.029$ \\
\hline
\textbf{DGM} & \textbf{Sampling} & \textbf{USR} & \textbf{OSR} & \textbf{Train f1} & \textbf{Test f1} \\
\hline
vae & Minority & $0.8$ & $0.9$ & $0.550 \pm 0.004$ & $0.535 \pm 0.030$ \\
mv-vae & Minority & $0.8$ & $0.9$ & $0.550 \pm 0.004$ & $0.533 \pm 0.031$ \\
arae & Minority & $0.7$ & $0.8$ & $0.546 \pm 0.004$ & $0.533 \pm 0.030$ \\
mv-arae & Minority & $0.8$ & $0.9$ & $0.548 \pm 0.005$ & $0.535 \pm 0.030$ \\
medgan & Minority & $0.8$ & $0.9$ & $0.546 \pm 0.005$ & $0.534 \pm 0.032$ \\
mv-medgan & Minority & $0.7$ & $0.8$ & $0.543 \pm 0.003$ & $0.534 \pm 0.032$ \\
gan & Minority & $0.8$ & $0.9$ & $0.549 \pm 0.004$ & $0.535 \pm 0.029$ \\
mv-wgan & Minority & $0.8$ & $0.9$ & $0.545 \pm 0.004$ & $0.534 \pm 0.030$ \\
mv-wgan-gp & Minority & $0.8$ & $0.9$ & $0.548 \pm 0.004$ & $0.534 \pm 0.031$ \\
\hline
vae & Conditional & $0.7$ & $0.9$ & $0.539 \pm 0.004$ & $0.531 \pm 0.030$ \\
mv-vae & Conditional & $0.7$ & $0.8$ & $0.546 \pm 0.005$ & $0.532 \pm 0.032$ \\
arae & Conditional & $0.8$ & $0.9$ & $0.549 \pm 0.006$ & $0.535 \pm 0.030$ \\
mv-arae & Conditional & $0.8$ & $0.9$ & $0.547 \pm 0.004$ & $0.533 \pm 0.030$ \\
medgan & Conditional & $0.8$ & $0.9$ & $0.548 \pm 0.003$ & $0.535 \pm 0.030$ \\
mv-medgan & Conditional & $0.8$ & $0.9$ & $0.546 \pm 0.004$ & $0.535 \pm 0.032$ \\
gan & Conditional & $0.8$ & $0.9$ & $0.550 \pm 0.003$ & $0.535 \pm 0.029$ \\
mv-wgan & Conditional & $0.8$ & $0.9$ & $0.546 \pm 0.004$ & $0.536 \pm 0.032$ \\
mv-wgan-gp & Conditional & $0.9$ & $1.0$ & $0.545 \pm 0.006$ & $0.534 \pm 0.028$ \\
\hline
vae & Rejection & $0.7$ & $0.8$ & $0.543 \pm 0.004$ & $0.532 \pm 0.031$ \\
mv-vae & Rejection & $0.7$ & $0.8$ & $0.545 \pm 0.004$ & $0.534 \pm 0.031$ \\
arae & Rejection & $0.9$ & $1.0$ & $0.546 \pm 0.004$ & $0.532 \pm 0.032$ \\
mv-arae & Rejection & $0.8$ & $0.9$ & $0.547 \pm 0.003$ & $0.534 \pm 0.032$ \\
medgan & Rejection & \multicolumn{4}{c|}{\textit{Timeout}} \\
mv-medgan & Rejection & \multicolumn{4}{c|}{\textit{Timeout}} \\
gan & Rejection & \multicolumn{4}{c|}{\textit{Timeout}} \\
mv-wgan & Rejection & $0.8$ & $0.9$ & $0.547 \pm 0.004$ & $0.535 \pm 0.032$ \\
mv-wgan-gp & Rejection & $0.9$ & $1.0$ & $0.547 \pm 0.004$ & $0.535 \pm 0.033$ \\
\hline
\end{tabular}
\end{table*}

\subsection{Datasets}

We select two datasets from the UCI Repository~\cite{dua_uci_2017}: ``Adult'' and ``Default of credit card clients''.
We also include the ``Credit card fraud'' dataset presented in \cite{dal2015calibrating}, which we obtained from the Kaggle repositories\footnote{https://www.kaggle.com/mlg-ulb/creditcardfraud}.
This dataset only contain numerical features (most of them from coming from a PCA transformation) and is highly imbalanced ($< 0.001\%$ of the cases are labeled as frauds).
In all three cases, we apply the same preprocessing procedure.
For all the categorical variables we use one-hot-encoding, but for the special case of binary variables, we just represent them with one binary feature.
Additionally, all the numerical variables are scaled to fit inside the range $[0; 1]$.
We generate metadata indicating for each variable the type (categorical, binary or numerical) and the size (number of features).
This information is used by the SMOTE-NC oversampling method, the autoencoder reconstruction loss and the multi-variable architectures.
The code for the preprocessing is available online~\footnote{https://github.com/rcamino/dataset-pre-processing}.

\subsection{Classification Protocol}

All of our experiments involve binary classification tasks implemented with XGBoost~\cite{Chen:2016:XST:2939672.2939785}.
We compute the mean and standard deviation of the f1 score for the train and test sets over 10 folds.
For each dataset, we run a grid search over several XGBoost hyperparameters (e.g. the max depth and number of estimators).
In the first section of Table~\ref{table:adult} and Table~\ref{table:default-of-credit-card-clients} we indicate the IR for each dataset along with the best results.
The selected XGBoost hyperparameters are fixed for the rest of the experiments that involve unsersampling and oversampling.

\subsection{Undersampling and Oversampling}

All the under- and oversampling algorithms presented in the experiments that do not involve deep generative models are taken from the imbalanced-learn library~\cite{JMLR:v18:16-365}.
Both datasets in the experiments are imbalanced, which means that the imbalance ratio (IR) is less than one, where IR is defined as:

\[
IR = \frac{|\{\text{minority class samples}\}|}{|\{\text{majority class samples}\}|}
\]

We undersample the majority class on the train set applying a random under sampler, which consists in removing samples from the majority class with a uniform probability, until reaching a desired sample size smaller than the original.
The result contains a larger IR that we call ``undersampling ratio'' (USR).
Afterwards, we train and evaluate the classification model and compute the respective metrics.
We repeat the process for incrementally larger values for the USR.
In the second section of Table~\ref{table:adult} and Table~\ref{table:default-of-credit-card-clients} we present the USR that provide the best mean test f1 score for each dataset.

Furthermore, for each USR, we oversample the minority class on the train set that was previously undersampled.
The result contains an even larger IR that we call ``oversampling ratio'' (OSR).
We repeat the process for incrementally larger values for the OSR and for different oversampling algorithms.
In the third section of Table~\ref{table:adult} and Table~\ref{table:default-of-credit-card-clients} we present the USR-OSR combination that provide the best mean test f1 score for each algorithm on each dataset.

\subsection{Oversampling with Deep Generative Models}

For the final experiments, we combine 9 deep generative models (DGM) with 3 different ways of sampling from them (presented in Section~\ref{sec:comparison}).
All the models are implemented with PyTorch~\cite{paszke_automatic_2017}.
The training of the models is done based on the training set, and the validation set is also used to evaluate the quality of the reconstruction when there is an autoencoder present in the architecture.
After the model training, the undersampling is applied to the training set, and afterwards the trained model generates synthetic samples that are appended to the training set based on the OSR.
In the fourth section of Table~\ref{table:adult} and Table~\ref{table:default-of-credit-card-clients} we present the USR-OSR combination that provide the best mean test f1 score for each dataset.
Note that the rejection sampling can ``timeout'' if it fails to obtain the desired amount of samples class after 10,000 iterations.

\subsection{Results}

We start analyzing the results shown on Table~\ref{table:creditcardfraud} for the dataset ``Credit card fraud''.
Given that the dataset contains no categorical or binary features, we do not implement the multi-variable models for this scenario.
The class imbalance is very high, and our experiments involving large changes to the imbalance ratio after under- and oversampling resulted in a large deterioration of the classification scores.
Small changes on the imbalance ratio after undersampling do not offer classification improvements either, but some cases of small oversampling appear to be useful.
In our experiments with this dataset, SMOTE presents the best classification score, while most of the deep generative models offer a slight improvement over the classification without oversampling.
Experiments with SMOTE and ``minority'' GAN were carried out with the same dataset in \cite{fiore2019using}, where both models presented very close results.
Besides the scores for SMOTE, our results are comparable to this study.
We believe that the small discrepancies are related to the stochastic nature of the experiments.
Additionally, we can see that the classification tends to overfit on every case, which is expected considering the small amount of positive samples.
Finally, note that all the models implementing rejection sampling reached a timeout.
This is expected for this dataset given that the probability of drawing a sample that belongs to the minority class is very low.

Now we compare all the results from Table~\ref{table:adult} and Table~\ref{table:default-of-credit-card-clients}.
Regarding the classification baseline without oversampling or undersampling, we can see that both problems are quite challenging, since the mean test f1 score is considerably far from 1.
Furthermore, adding only random undersampling to the pipeline shows some improvement in Table~\ref{table:adult} but presents a considerable improvement in Table~\ref{table:default-of-credit-card-clients}.
Nevertheless, most of the the oversampling techniques (both with and without deep generative models) only show a slight improvement on the third decimal of the mean test f1 score on top of the undersampling.
The only oversampling algorithm that seems to worsen the classification quality is SMOTE-NC.
Note also that in Table~\ref{table:default-of-credit-card-clients} the RandomOverSampler --which is the simplest oversampling technique-- performs slightly better than the rest.

\begin{figure*}[!t]
\centering
\begin{minipage}{.45\textwidth}
  \centering
  \includegraphics[width=\textwidth]{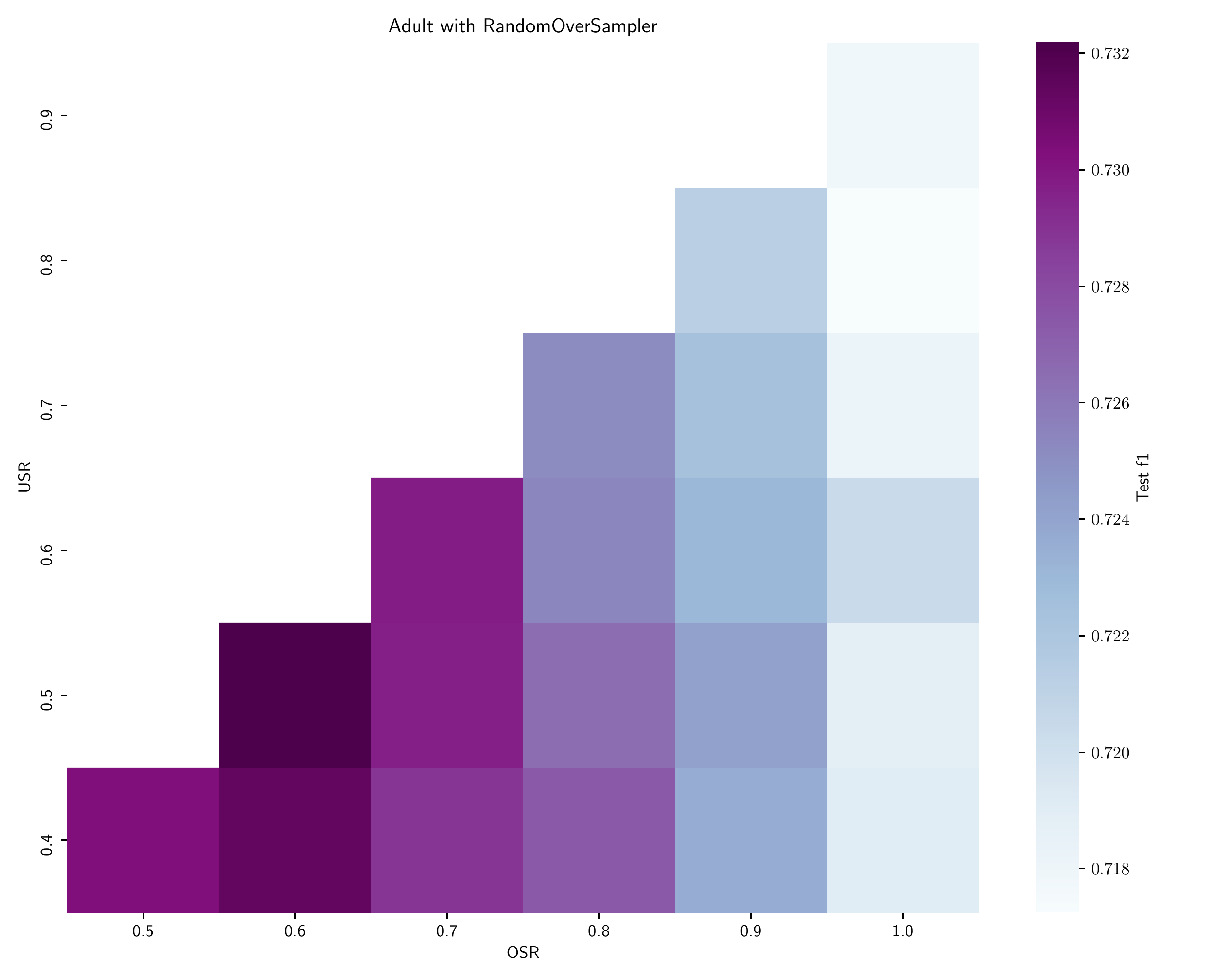}
\end{minipage}%
\begin{minipage}{.45\textwidth}
  \centering
  \includegraphics[width=\textwidth]{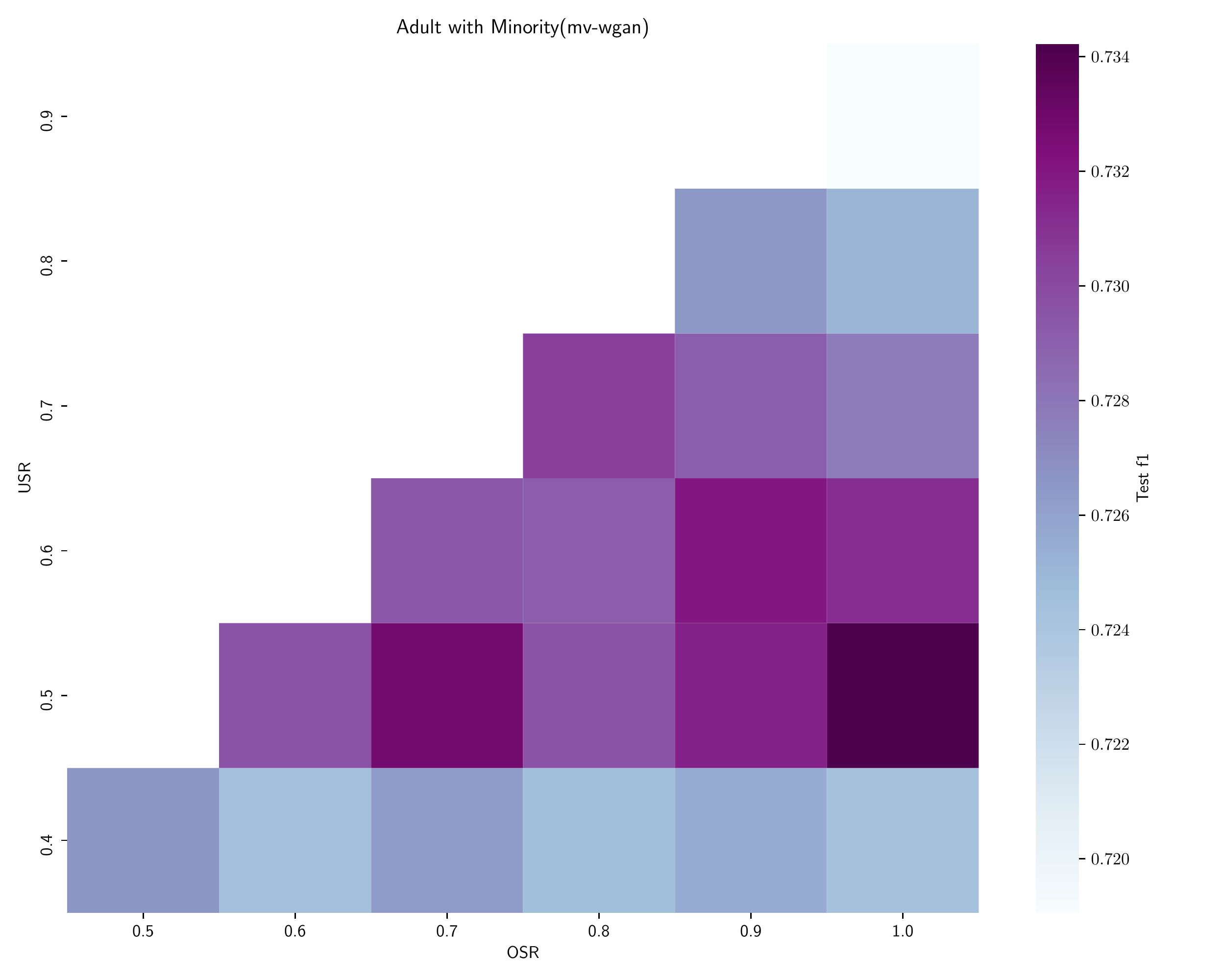}
\end{minipage}
\caption{Mean test f1 score for classifications using the ``Adult'' dataset, crossed with different values for USR and OSR. On the left the oversampling is implemented by a RandomOverSampler, and on the right, by drawing samples from a Multi-Variable WGAN trained only with the minority class.}
\label{fig:oversampling}
\end{figure*}

In Fig.~\ref{fig:oversampling} we present two example distributions of the f1 score across different ratios.
We can see that there is some difference between the oversampling with and without deep generative models.
The RandomOverSampler deteriorates progressively by incrementing the oversampling ratio, while the oversampling using a Multi-Variable WGAN does the opposite.
Nevertheless, note that the scale of the color bar (representing the mean test f1 score) is not very significant.

\begin{figure*}[!t]
\centering
\begin{minipage}{.30\textwidth}
  \centering
  \includegraphics[width=\textwidth]{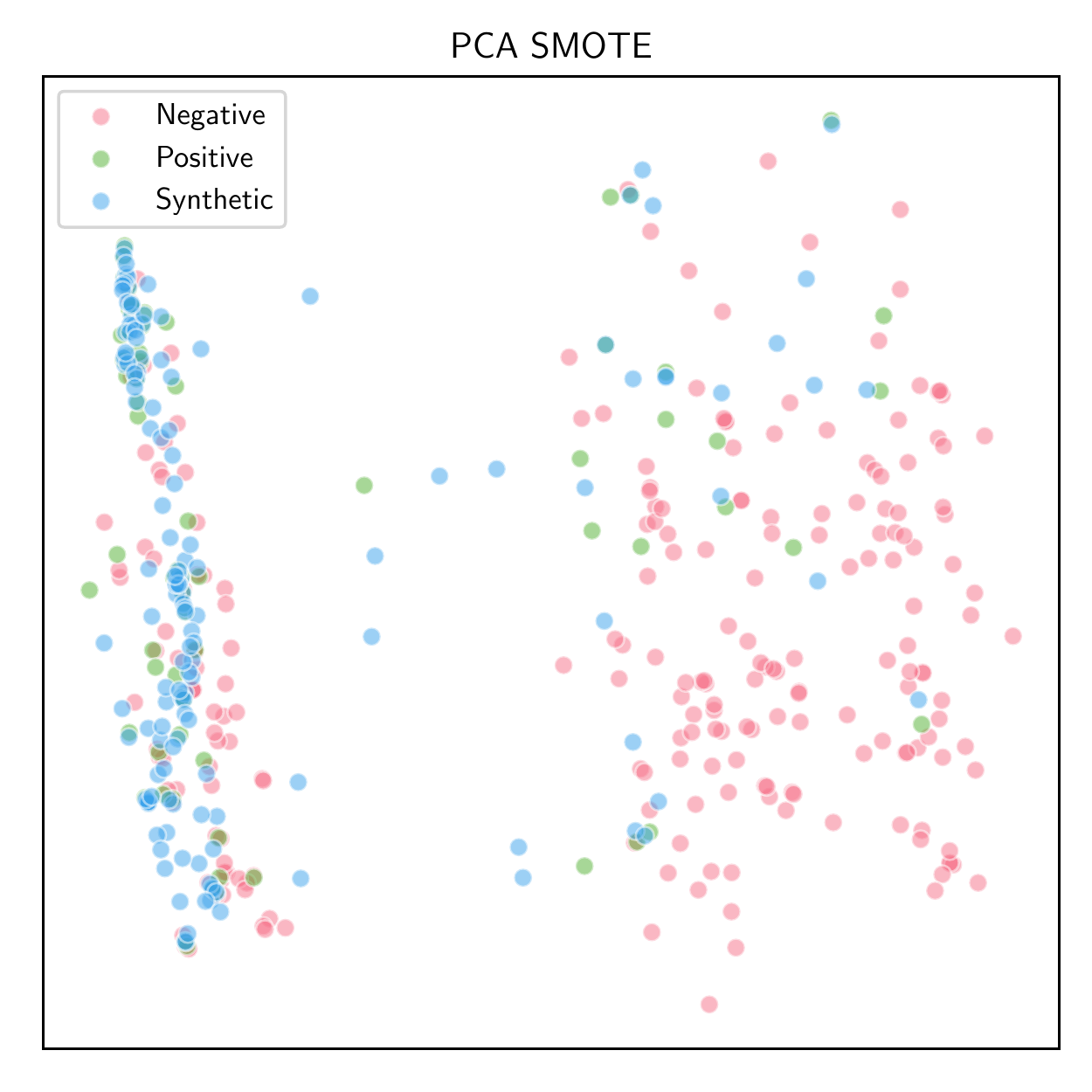}
\end{minipage}%
\begin{minipage}{.30\textwidth}
  \centering
  \includegraphics[width=\textwidth]{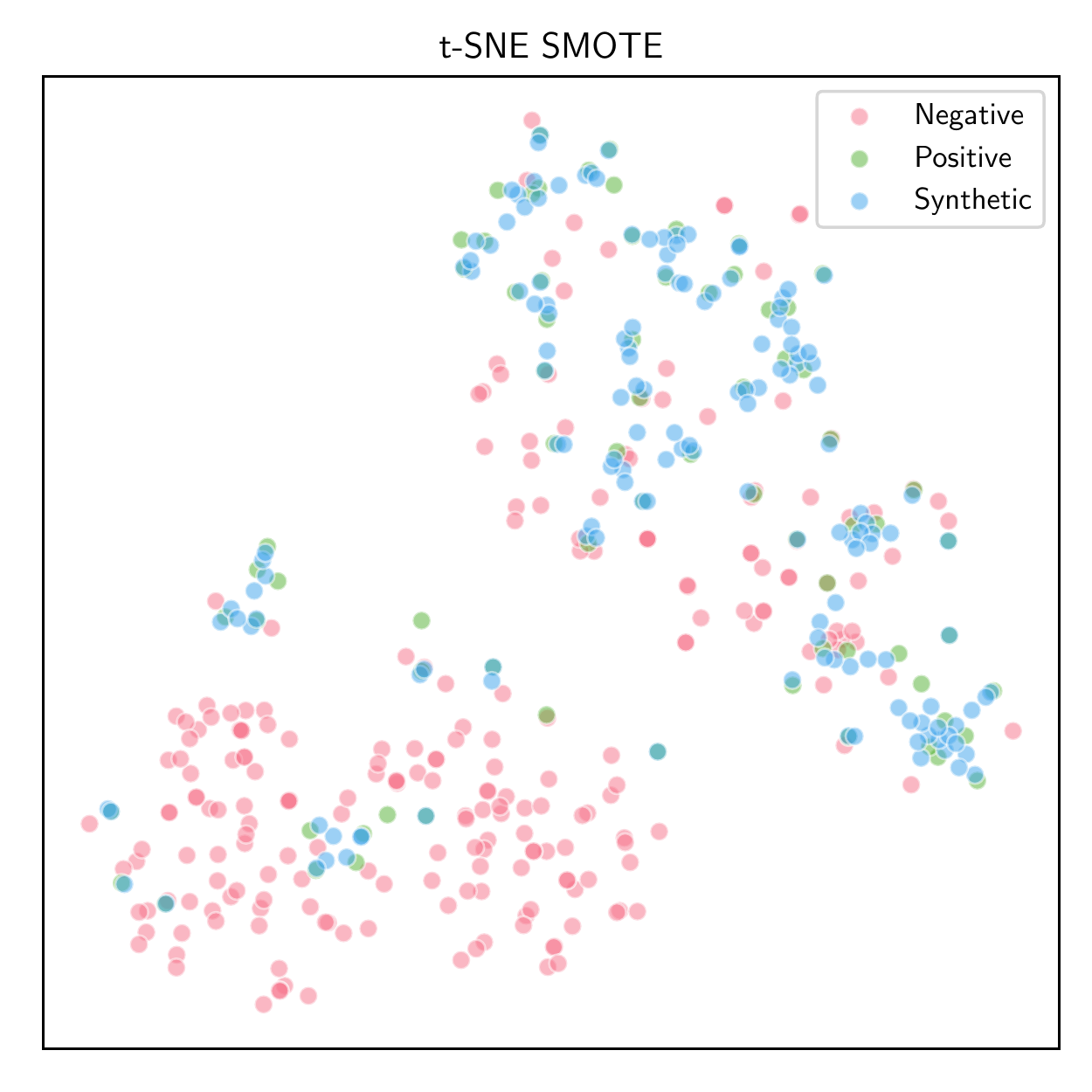}
\end{minipage}%
\begin{minipage}{.30\textwidth}
  \centering
  \includegraphics[width=\textwidth]{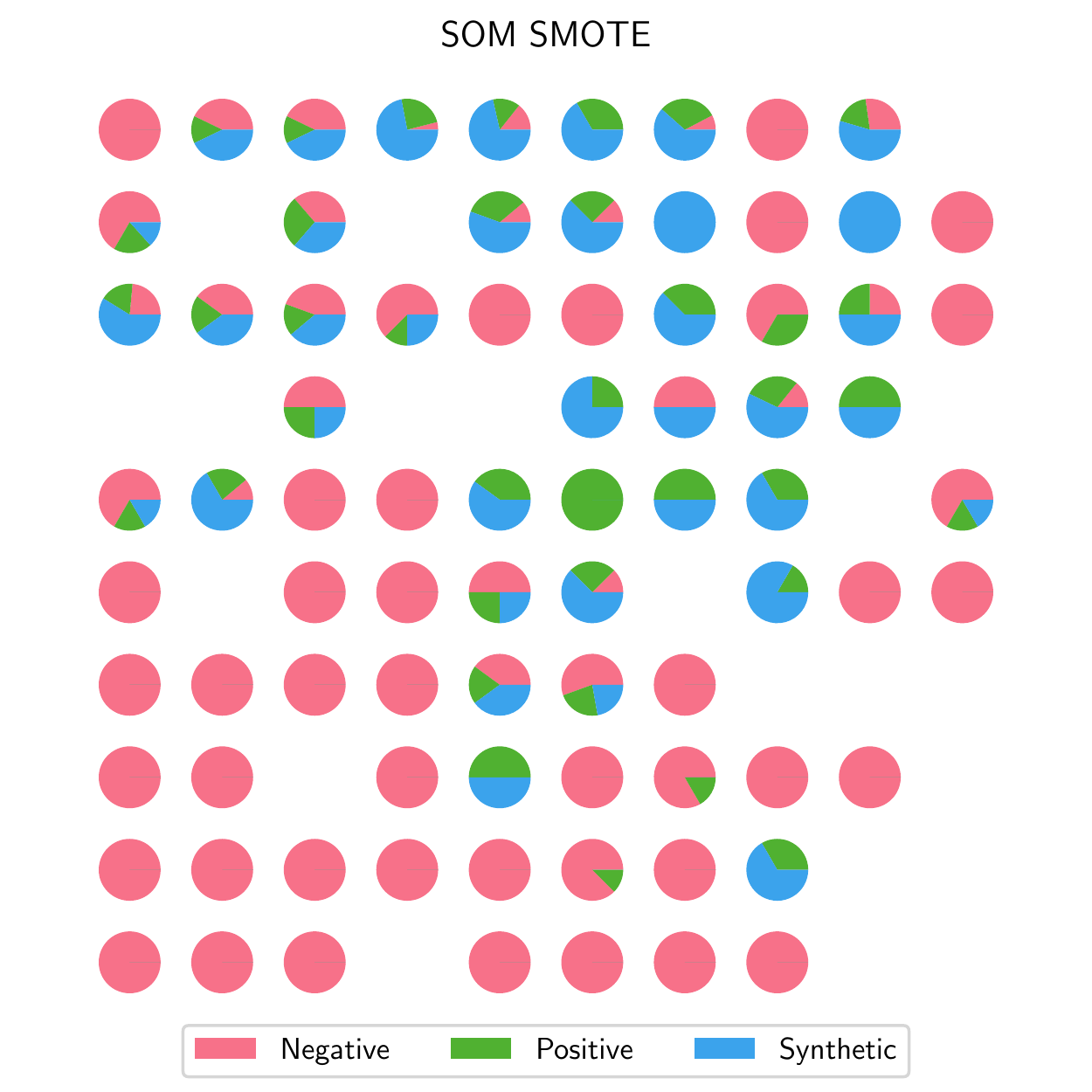}
\end{minipage}%
\label{fig:3plot-smote}
\end{figure*}

\begin{figure*}[!t]
\centering
\begin{minipage}{.30\textwidth}
  \centering
  \includegraphics[width=\textwidth]{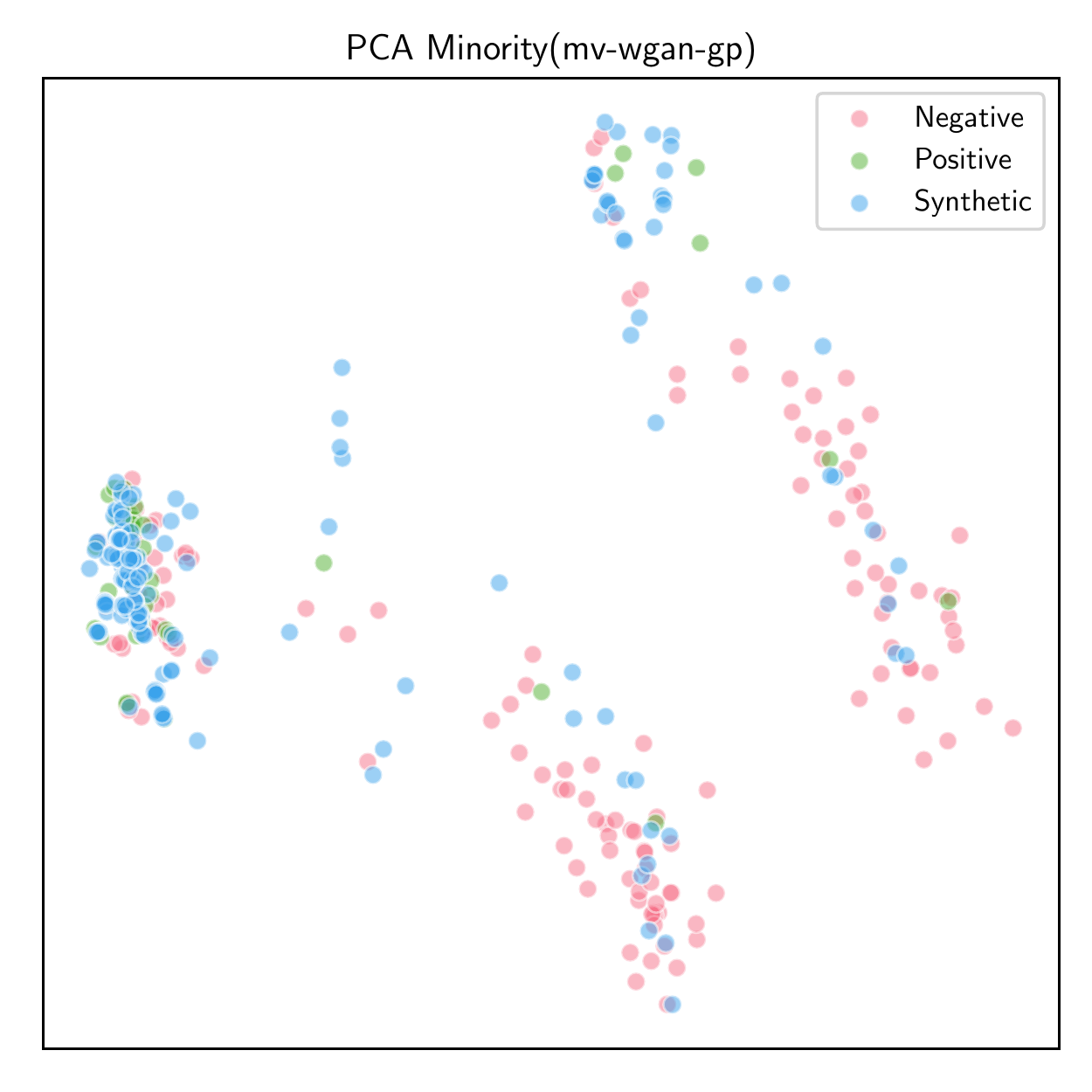}
\end{minipage}%
\begin{minipage}{.30\textwidth}
  \centering
  \includegraphics[width=\textwidth]{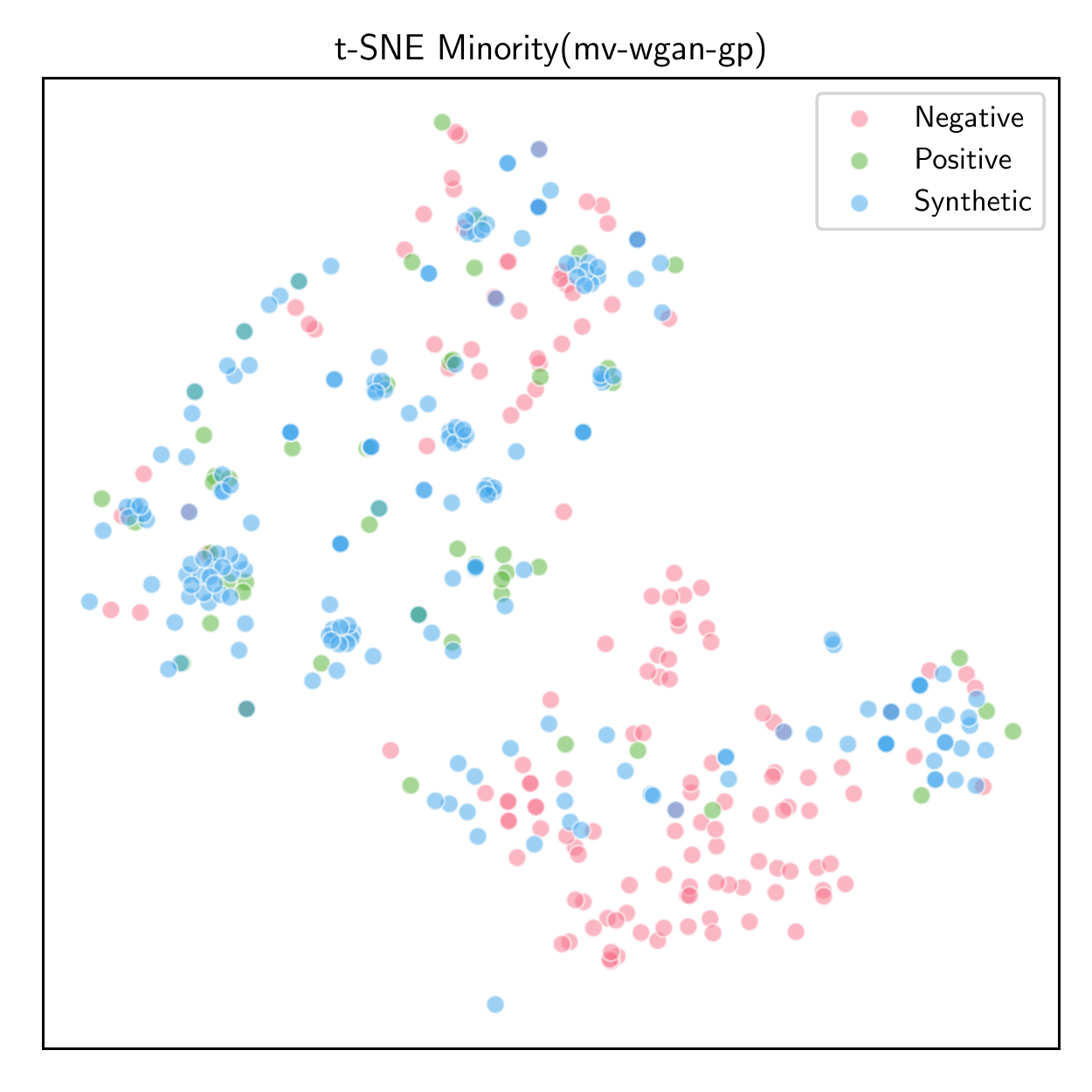}
\end{minipage}%
\begin{minipage}{.30\textwidth}
  \centering
  \includegraphics[width=\textwidth]{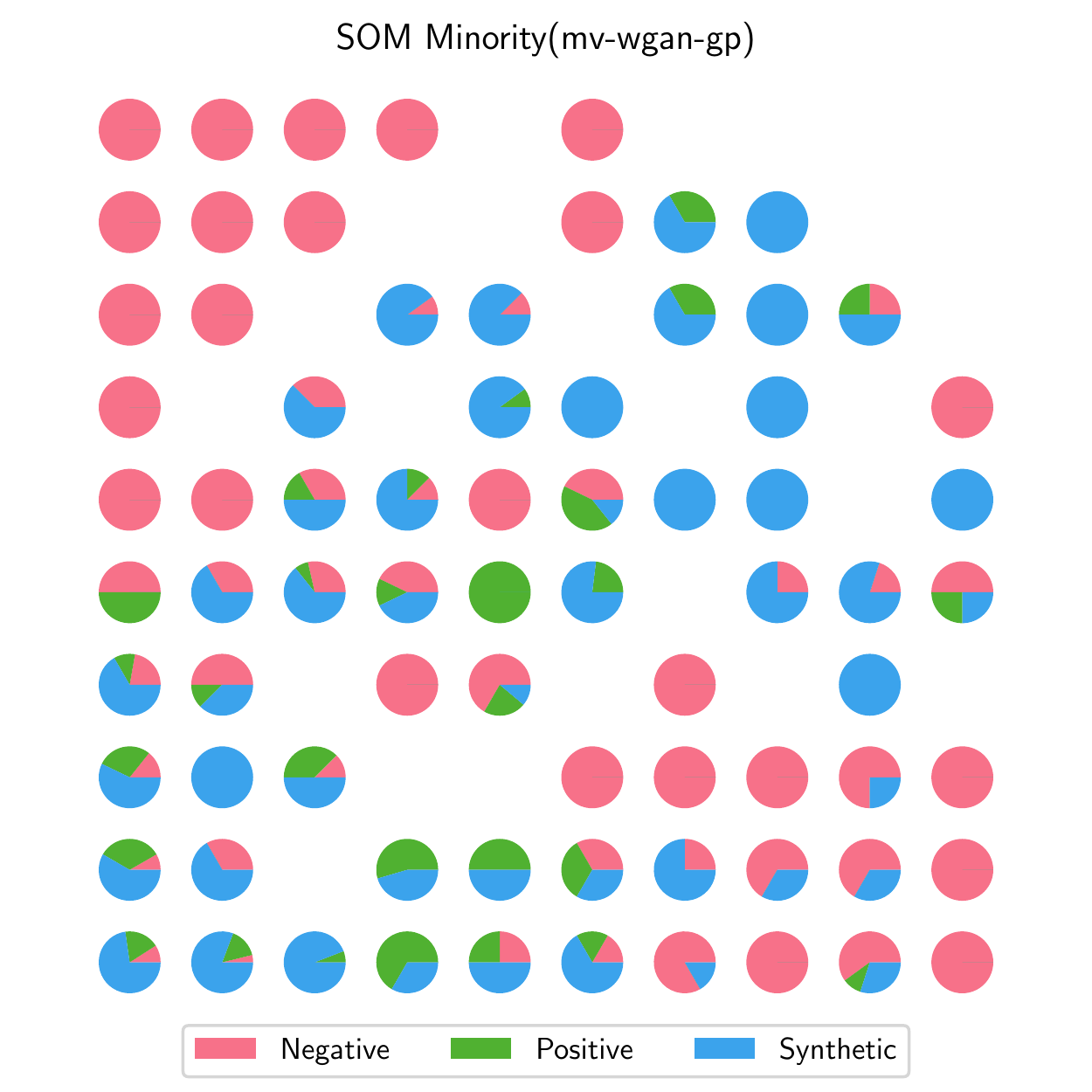}
\end{minipage}%
\label{fig:3plot-minority-mv-wgan}
\end{figure*}

\begin{figure*}[!t]
\centering
\begin{minipage}{.30\textwidth}
  \centering
  \includegraphics[width=\textwidth]{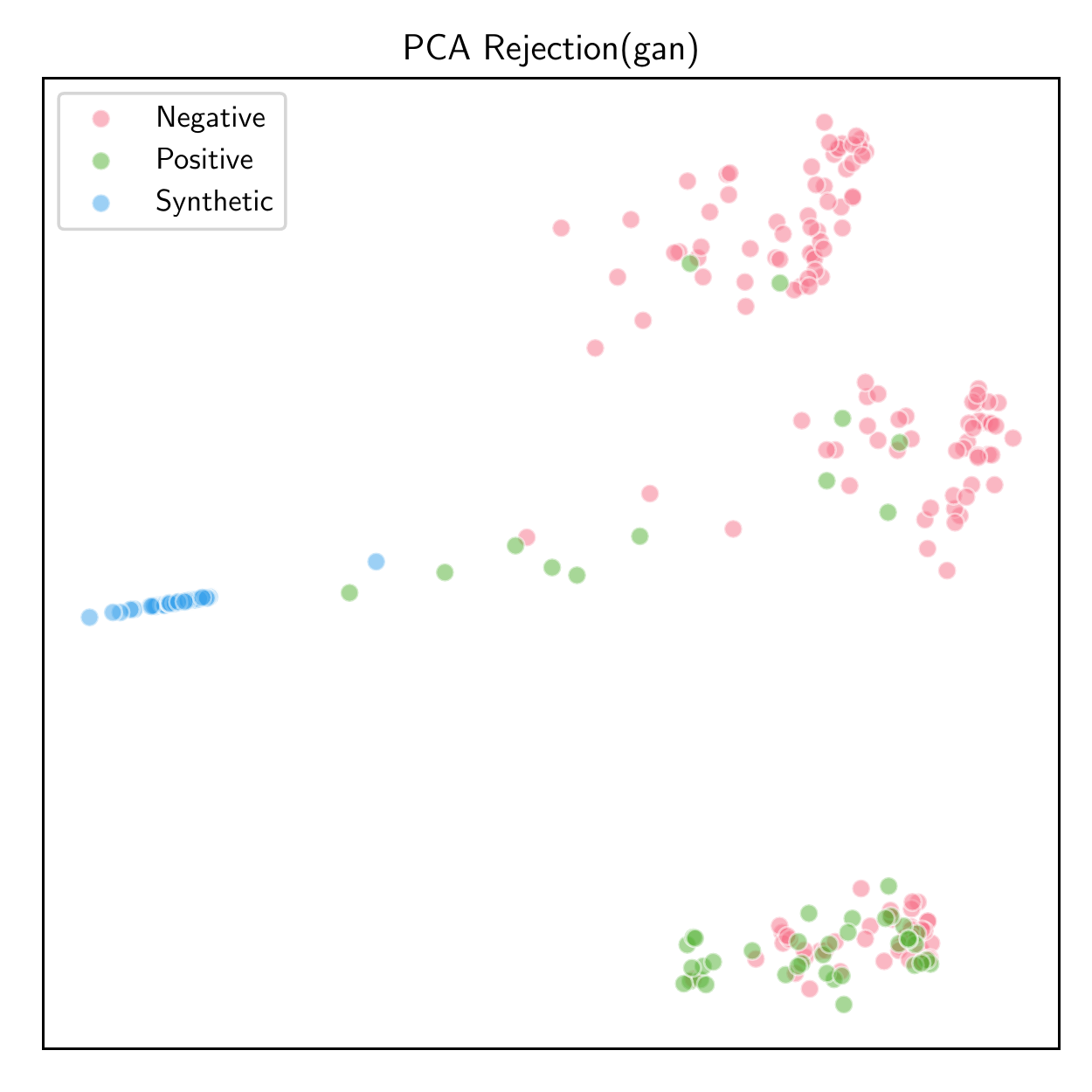}
\end{minipage}%
\begin{minipage}{.30\textwidth}
  \centering
  \includegraphics[width=\textwidth]{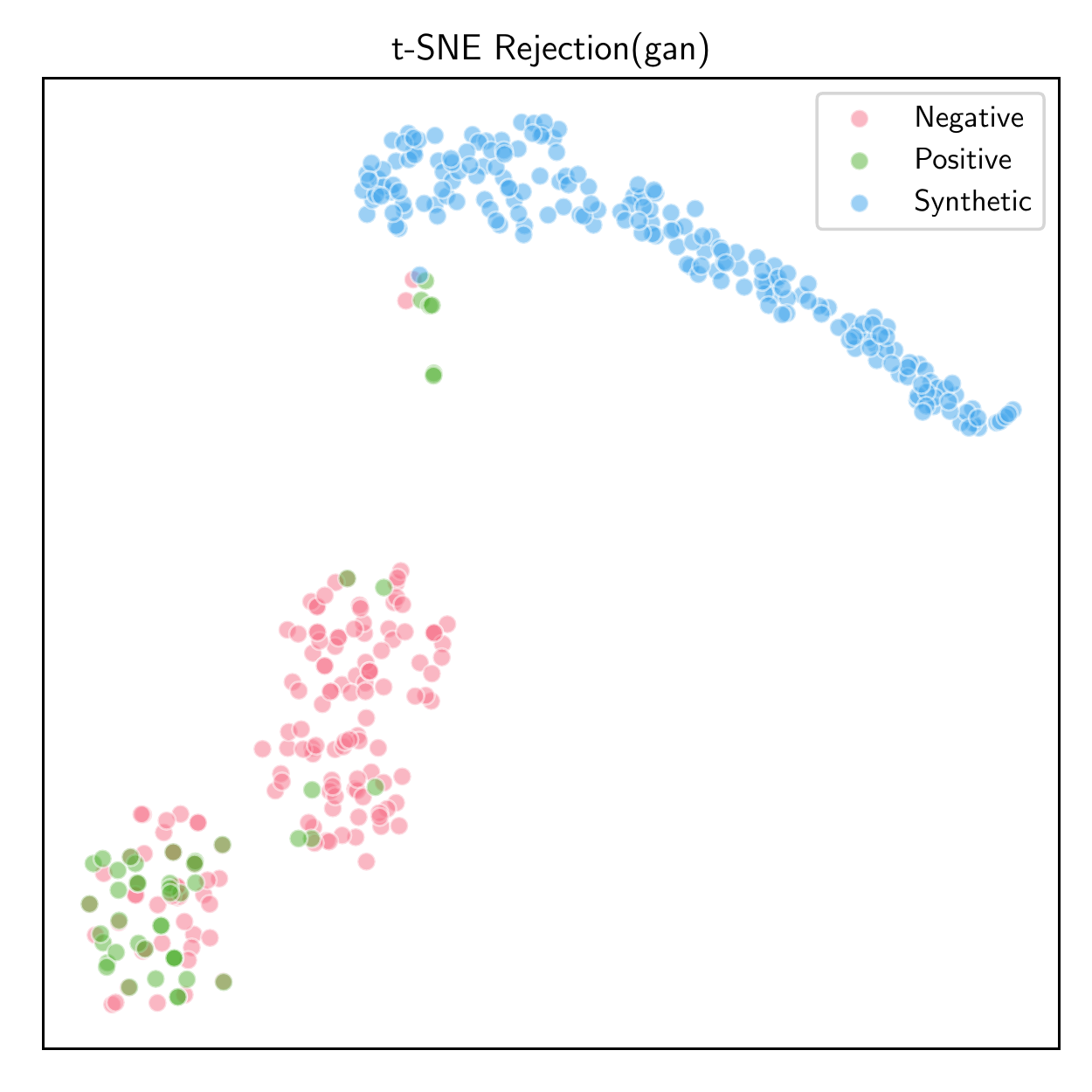}
\end{minipage}%
\begin{minipage}{.30\textwidth}
  \centering
  \includegraphics[width=\textwidth]{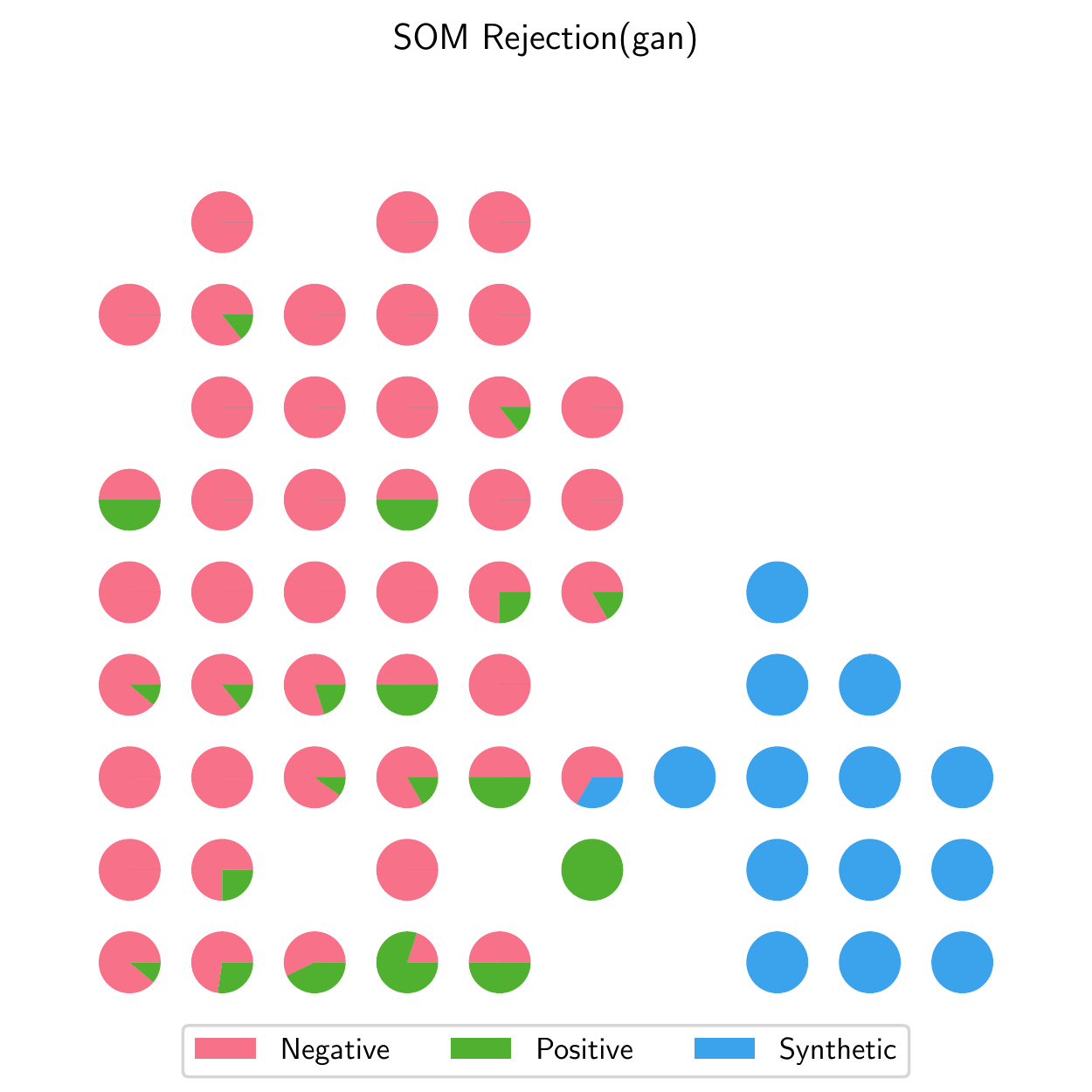}
\end{minipage}%
\caption{Nine different visualization examples built from 200 datapoints drawn at random from the ``Adult'' dataset and augmented with 200 synthetic datapoints. On each colum there is a different visualization: PCA and t-SNE transformations into two dimensions followed by a 10x10 SOM. On each row a different oversampling technique: SMOTE, MV-WGAN-GP with minority sampling and GAN with rejection sampling.}
\label{fig:3plot-rejection-gan}
\end{figure*}

In Fig.~\ref{fig:3plot-rejection-gan} we intent so explore visually some of the differences between sampling techniques.
We present three alternatives: two feature space transformations into two dimensions based on PCA and t-SNE \cite{maaten2008visualizing}, and Self Organizing Maps (SOM) \cite{kohonen1990self}.
The transformations are presented in scatter plots where samples are colored as negative, positive or synthetic (drawn from an oversampling algorithm).
These plots provide us with the tools to examine how close to each other are samples from different types.
On the other side, the SOM assigns samples to different cells on a grid, where we compare the frequency of each sample type with a pie chart.
Given that these methods do not scale well with the number of samples, we only visualize a subset in each plot.
Note that we are losing information by dropping samples and dimensions at the same time, but these approximations can still give us an intuition of how the data is organized.
In the first two rows we show SMOTE and one of the DGM models, and in both cases we can see that synthetic samples tend to be grouped with positive cases as desired.
Also, many of the negative classes are far away from the other two types of samples.
Furthermore, in the third row, we show one example of a DGM model that generates samples that are very separated from the rest of the sample types.
Nevertheless, regardless of this two different situations, we see in the classification results of the corresponding Table \ref{table:adult} that all oversampling techniques achieve practically the same performance.
This could mean that the classification boundaries computed by XGBoost do not change much neither by adding synthetic samples resembling positive samples, nor with synthetic samples that fill empty portions of the feature space.
\section{Conclusion}
\label{sec:conclusion}

Our experiments show several trends that persist across different generative methods and datasets:
First, undersampling the majority class without any oversampling improves the classifier.
Adding oversampling via a simple baseline such as SMOTE only leads to marginal improvements.
Second, all generative models provide an improvement over the baselines.
The sampling strategy does not seem to have a strong impact on the quality of the results, but rejection sampling clearly is the slowest approach (it may run indefinitely).

It is noteworthy that generative models require different under- and oversampling settings to archive best performance (c.f.  Fig.~\ref{fig:oversampling}).
Yet, regardless of the method used, the absolute improvement on the classification metric (F1 score) is often very small. 
This is despite results showing that in a ranking of methods the improvement of deep generative methods is statistically significantly better~\cite{douzas2018effective}.

It seems that the performance gain given by deep generative models for oversampling has to be seen in context.
They posses a considerably more complicated set-up and longer training time when compared to the best-performing baseline, a simple random under- and oversampling approach.






\bibliographystyle{IEEEtran}
\bibliography{Zotero}
%



\end{document}